\pdfoutput=1

\documentclass[final]{cvpr}

\usepackage{times}
\usepackage{epsfig}
\usepackage{graphicx}
\usepackage{amsmath}
\usepackage{amssymb}
\usepackage{comment}

\usepackage[pagebackref=true,breaklinks=true,colorlinks,bookmarks=false]{hyperref}

\def\cvprPaperID{10441} 
\def\confYear{CVPR 2021}

\begin{document}

\title{Dynamic Head: Unifying Object Detection Heads with Attentions}

\author{Xiyang Dai \quad Yinpeng Chen \quad Bin Xiao \quad Dongdong Chen \quad Mengchen Liu \\
\quad Lu Yuan \quad Lei Zhang\\
Microsoft\\
Redmond, USA\\
{\tt\small \{xidai, yiche, bixi, dochen, mengcliu, luyuan, leizhang\}@microsoft.com}
}

\maketitle

\begin{abstract}
The complex nature of combining localization and classification in object detection has resulted in the flourished development of methods. Previous works tried to improve the performance in various object detection heads but failed to present a unified view. In this paper, we present a novel dynamic head framework to unify object detection heads with attentions. By coherently combining multiple self-attention mechanisms between feature levels for scale-awareness, among spatial locations for spatial-awareness, and within output channels for task-awareness, the proposed approach significantly improves the representation ability of object detection heads without any computational overhead. Further experiments demonstrate that the effectiveness and efficiency of the proposed dynamic head on the COCO benchmark. With a standard ResNeXt-101-DCN backbone, we largely improve the performance over popular object detectors and achieve a new state-of-the-art at 54.0 AP. Furthermore, with latest transformer backbone and extra data, we can push current best COCO result to a new record at 60.6 AP. The code will be released at \url{https://github.com/microsoft/DynamicHead}.

\end{abstract}

\section{Introduction}

Object detection is to answer the question ``what objects are located at where" in computer vision applications. In the deep learning era, nearly all modern object detectors~\cite{fastrcnn, fasterrcnn, maskrcnn, atss, fcos, spec, reppoints} share the same paradigm -- a backbone for feature extraction and a head for localization and classification tasks. How to improve the performance of an object detection head has become a critical problem in existing object detection works. 

The challenges in developing a good object detection head can be summarized into three categories. Firstly, the head should be \textit{scale-aware}, since multiple objects with vastly distinct scales often co-exist in an image. Secondly, the head should be \textit{spatial-aware}, since objects usually appear in vastly different shapes, rotations, and locations under different viewpoints. Thirdly, the head needs to be \textit{task-aware}, since objects can have various  representations (\eg, bounding box~\cite{maskrcnn}, center~\cite{fcos}, and corner points~\cite{reppoints}) that own totally different objectives and constraints. We find recent studies \cite{maskrcnn, atss, fcos, spec, reppoints} only focus on solving one of the aforementioned problems in various ways. It remains an open problem how to develop a unified head that can address all these problems simultaneously.

In this paper, we propose a novel detection head, called \textit{dynamic head}, to unify scale-awareness, spatial-awareness, and task-awareness all together. If we consider the output of a backbone (\ie, the input to a detection head) as a 3-dimensional tensor with dimensions $ level \times space \times channel$, we discover that such a unified head can be regarded as an attention learning problem. An intuitive solution is to build a full self-attention mechanism over this tensor. However, the optimization problem would be too difficult to solve and the computational cost is not affordable. 

Instead, we can deploy attention mechanisms separately on each particular dimension of features, \ie, level-wise, spatial-wise, and channel-wise. The scale-aware attention module is only deployed on the dimension of $level$. It learns the relative importance of various semantic levels to enhance the feature at a proper level for an individual object based on its scale. The spatial-aware attention module is deployed on the dimension of $space$ (\ie, $height \times width$). It learns coherently discriminative representations in spatial locations. The task-aware attention module is deployed on $channels$. It directs different feature channels to favor different tasks separately (\eg, classification, box regression, and center/key-point learning.) based on different convolutional kernel responses from objects.   

In this way, we explicitly implement a unified attention mechanism for the detection head. Although these attention mechanisms are separately applied on different dimensions of a feature tensor, their performance can complement each other. Extensive experiments on the MS-COCO benchmark demonstrate the effectiveness of our approach. It offers a great potential for learning a better representation that can be utilized to improve all kinds of object detection models with $1.2\% \sim 3.2\%$ AP gains. With the standard ResNeXt-101-DCN backbone, the proposed method achieves a new state of the art $54.0\%$ AP on COCO. Besides, compared with EffcientDet~\cite{efficientdet} and SpineNet~\cite{spinenet}, dynamic head uses $1/20$  training time, yet with a better performance. Furthermore, with latest transformer backbone and extra data from self-training, we can push current best COCO result to a new record at 60.6 AP (see appendix for details).

\section{Related Work}

Recent studies focus on improving object detectors from various perspectives: scale-awareness, spatial-awareness and task-awareness.  
\paragraph{Scale-awareness.} 
Many researches have empathized the importance of scale-awareness in object detection as objects with vastly different scales often co-exist in natural images. Early works have demonstrated the significance of leveraging image pyramid methods \cite{rcnn, snip, sniper} for multi-scale training. 
Instead of image pyramid, feature pyramid \cite{fpn} was proposed to improve efficiency by concatenating a pyramid of down-sampled convolution features and had become a standard component in modern object detectors. However, features from different levels are usually extracted from different depth of a network, which causes a noticeable semantics gap. To solve this discrepancy, \cite{panet} proposed to enhance the features in lower layers by bottom-up path augmentation from feature pyramid. Later, \cite{libra} improved it by introducing balanced sampling and balanced feature pyramid. Recently, \cite{spec} proposed a pyramid convolution to extract scale and spatial features simultaneously based on a modified 3-D convolution.

In this work, we present a scale-aware attention in the detection head, which makes the importance of various feature level adaptive to the input. 

\paragraph{Spatial-awareness.}
Previous works have tried to improve the spatial-awareness in object detection for better semantic learning. Convolution neural networks were known to be limited in learning spatial transformations existed in images \cite{od-20y}. Some works mitigate this problem by either increasing the model capability (size) \cite{resnet, resnext} or involving expensive data augmentations \cite{aug}, resulting in extremely high computational cost in inference and training. Later, new convolution operators were proposed to improve the learning of spatial transformations. \cite{dilated} proposed to use dilated convolutions to aggregate contextual information from the exponentially expanded receptive
field. \cite{deform} proposed a deformable convolution to sample spatial locations with additional self-learned offsets. \cite{deformv2} reformulated the offset by introducing a learned feature amplitude and further improved its ability.

In this work, we present a spatial-aware attention in the detection head, which not only applies attention to each spatial location, but also adaptively aggregates multiple feature levels together for learning a more discriminative representation. 

\paragraph{Task-awareness.}
Object detection was originated from a two-stage paradigm \cite{edgebox, rcnn}, which first generates object proposals and then classifies the proposals into different classes and background. \cite{fasterrcnn} formalized the modern two-stage framework by introducing Region Proposal Networks (RPN) to formulate both stages into a single convolution network. Later, one-stage object detector \cite{yolo} became popular due to its high efficiency. \cite{retinanet} further improved the architecture by introducing task-specific branches to surpass the accuracy of two-stage detectors while maintaining the speed of previous one-stage detectors. 

Recently, more works have discovered that various representations of objects could potentially improve the performance.
\cite{maskrcnn} first demonstrated that combining bounding box and segmentation mask of objects can further improve the performance. \cite{fcos} proposed to use center representations to solve object detection in a per-pixel prediction fashion. \cite{atss} further improved the performance of center-based method by automatically selecting positive and negative samples according to statistical characteristics of object. Later, \cite{reppoints} formulated object detection as representative key-points to ease the learning. \cite{centernet} further improved the performance by detecting each object as a triplet, rather than a pair of key-points to reduce incorrect predictions. Most recently, \cite{borderdet} proposed to extract border features from the extreme points of each border to enhance the point feature and archived the state-of-the-art performance. 

In this work, we present a task-aware attention in the detection head, which allows attention to be deployed on channels, which can adaptively favor various tasks, for either single-/two-stage detectors, or box-/center-/keypoint-based detectors.

\paragraph{} More importantly, all the above properties are integrated into a unified attention mechanism in our head design. To our best knowledge, it is the first general detection head framework which takes a step towards understanding what role attention plays in the success of object detection head.

\section{Our Approach}

\begin{figure*}[t]
	\begin{center}
		\includegraphics[width=0.9\linewidth]{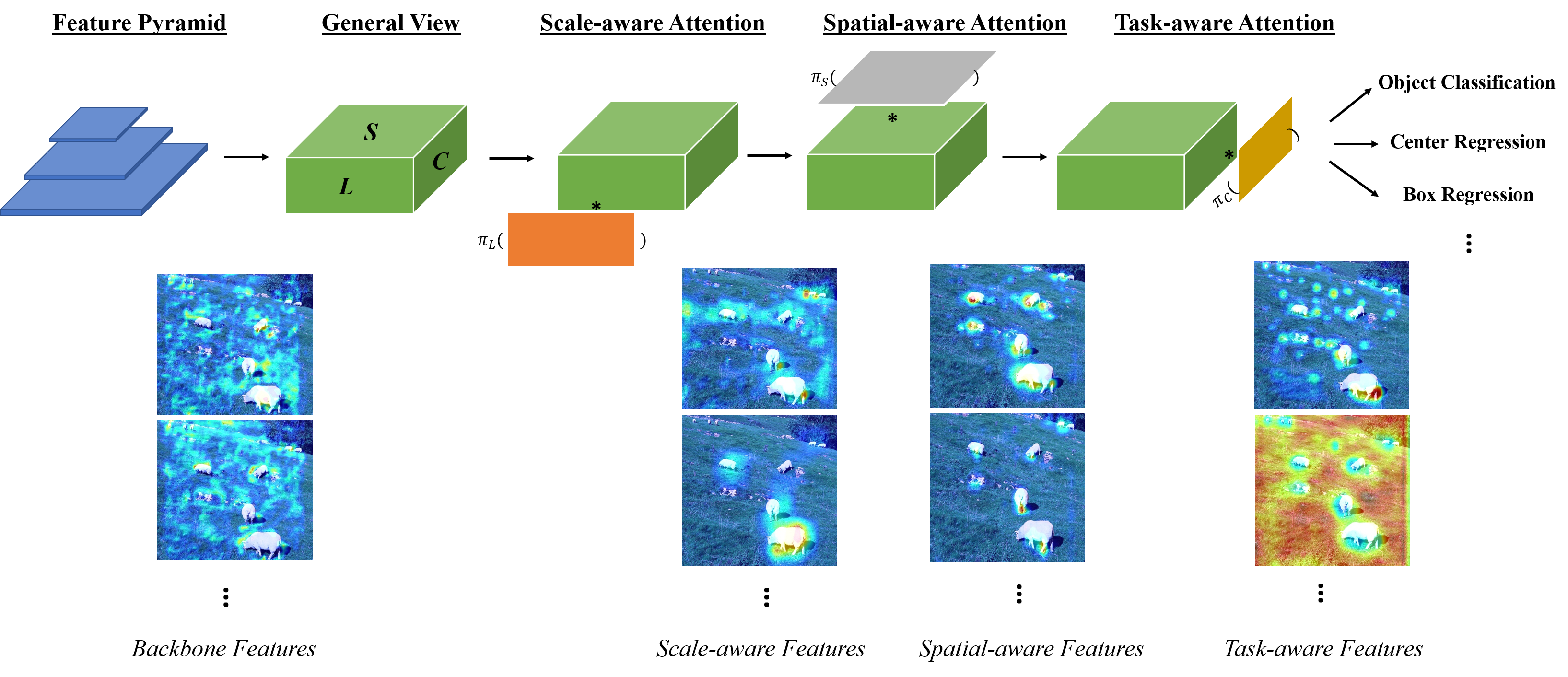}
	\end{center}
	\caption{An illustration of our Dynamic Head approach. It contains three different attention mechanisms, each focusing on a different perspective: scale-aware attention, spatial-aware attention, and task-aware attention. We also visualize how the feature maps are improved after each attention module. }
	\label{fig:main}
\end{figure*}

\subsection{Motivation}
In order to enable scale-awareness, spatial-awareness and task-awareness simultaneously in a unified object detection head, we need to generally understand previous improvements on object detection heads. 

Given a concatenation of features $\mathcal{F}_{in} = \{F_{i}\}_{i=1}^{L}$ from $L$ different levels in a feature pyramid, we can resize consecutive level features towards the scale of the median level feature using up-sampling or down-sampling. The re-scaled feature pyramid can be denoted as a 4-dimensional tensor $\mathcal{F}\in \mathcal{R}^{L\times H\times W\times C}$, where $L$ represents the number of levels in the pyramid, $H$, $W$, and $C$ represent height, width, and the number of channels of the median level feature respectively. We further define $S=H\times W$ to reshape the tensor into a 3-dimensional tensor $\mathcal{F}\in \mathcal{R}^{L\times S\times C}$. Based on this representation, we will explore the role of each tensor dimension. \vspace{-1.0em}
\begin{itemize}
    \item The discrepancy of object scales is related to features at various levels. Improving the representation learning across different levels of $\mathcal{F}$ can benefit scale-awareness of object detection.\vspace{-0.5em}
    \item Various geometric transformations from dissimilar object shapes are related to features at various spatial locations. Improving the representation learning across different spatial locations of $\mathcal{F}$ can benefit spatial-awareness of object detection.\vspace{-0.5em}
    \item Divergent object representations and tasks can be related to the features at various channels. Improving the representation learning across different channels of $\mathcal{F}$ can benefit task-awareness of object detection.
\end{itemize} 



In this paper, we discover that all above directions can be unified in an efficient attention learning problem. Our work is the first attempt to combine multiple attentions on all three dimensions to formulate a unified head for maximizing their improvements.

\subsection{Dynamic Head: Unifying with Attentions}
Given the feature tensor $\mathcal{F}\in \mathcal{R}^{ L\times S\times C}$, the general formulation of applying self-attention is:
\begin{equation}
    W(\mathcal{F}) = \pi(\mathcal{F})\cdot\mathcal{F}
\end{equation}
where $\pi(\cdot)$ is an attention function. A na\"ive solution to this attention function is implemented by fully connected layers. But directly learning the attention function over all dimensions is computationally costly and practically not affordable due to the high dimensions of the tensor.

Instead, we convert the attention function into three sequential attentions, each focusing on only one perspective:
\begin{equation}
   W(\mathcal{F}) = \pi_{C}\Bigg(\pi_{S}\bigg(\pi_{L}(\mathcal{F})\cdot\mathcal{F}\bigg)\cdot\mathcal{F}\Bigg)\cdot\mathcal{F},
\label{eq:main}
\end{equation}
where $\pi_{L}(\cdot)$, $\pi_{S}(\cdot)$, and $\pi_{C}(\cdot)$ are three different attention functions applying on dimension $L$, $S$, and $C$, respectively.

\paragraph{Scale-aware Attention $\pi_{L}$.}
We first introduce a scale-aware attention to dynamically fuse features of different scales based on their semantic importance. 
\begin{equation}
    \pi_{L}(\mathcal{F})\cdot\mathcal{F} = \sigma\Bigg(f\bigg(\frac{1}{SC}\sum_{S,C}\mathcal{F}\bigg)\Bigg)\cdot\mathcal{F}
\end{equation}
where $f(\cdot)$ is a linear function approximated by a $1\times1$ convolutional layer, and $\sigma(x) = max(0,min(1,\frac{x+1}{2}))$ is a hard-sigmoid function.

\paragraph{Spatial-aware Attention $\pi_{S}$.} We apply another spatial-aware attention module based on the fused feature to focus on discriminative regions consistently co-existing among both spatial locations and feature levels. Considering the high dimensionality in $S$, we decompose this module into two steps: first making the attention learning sparse by using deformable convolution~\cite{deform} and then aggregating features across levels at the same spatial locations:
\begin{equation}
    \pi_{S}(\mathcal{F})\cdot\mathcal{F} = \frac{1}{L}\sum_{l=1}^L\sum_{k=1}^K w_{l,k} \cdot \mathcal{F}(l; p_k+\Delta p_k; c) \cdot \Delta m_k,
\end{equation}
where $K$ is the number of sparse sampling locations,  $p_k+\Delta p_k$ is a shifted location by the self-learned spatial offset $\Delta p_k$ to focus on a discriminative region and $\Delta m_k$ is a self-learned importance scalar at location $p_k$. Both are learned from the input feature from the median level of $\mathcal{F}$.

\paragraph{Task-aware Attention $\pi_{C}$.} To enable joint learning and generalize different representations of objects, we deploy a task-aware attention at the end. It dynamically switches ON and OFF channels of features to favor different tasks: 

\begin{equation}
    \pi_{C}(\mathcal{F})\cdot\mathcal{F} = max\bigg(\alpha^1(\mathcal{F})\cdot\mathcal{F}_{c} + \beta^1(\mathcal{F}), \alpha^2(\mathcal{F})\cdot\mathcal{F}_{c} + \beta^2(\mathcal{F})\bigg),
\end{equation}
where $\mathcal{F}_{c}$ is the feature slice at the $c$-th channel and $[\alpha^1, \alpha^2, \beta^1, \beta^2]^T=\theta(\cdot)$ is a hyper function that learns to control the activation thresholds. $\theta(\cdot)$ is implemented similar to~\cite{dyrelu}, which first conducts a global average pooling on $L\times S$ dimensions to reduce the dimensionality, then uses two fully connected layers and a normalization layer, and finally applies a shifted sigmoid function to normalize the output to $[-1, 1]$.  

\begin{figure}[t]
	\begin{center}
	\hbox{\includegraphics[width=1.0\linewidth]{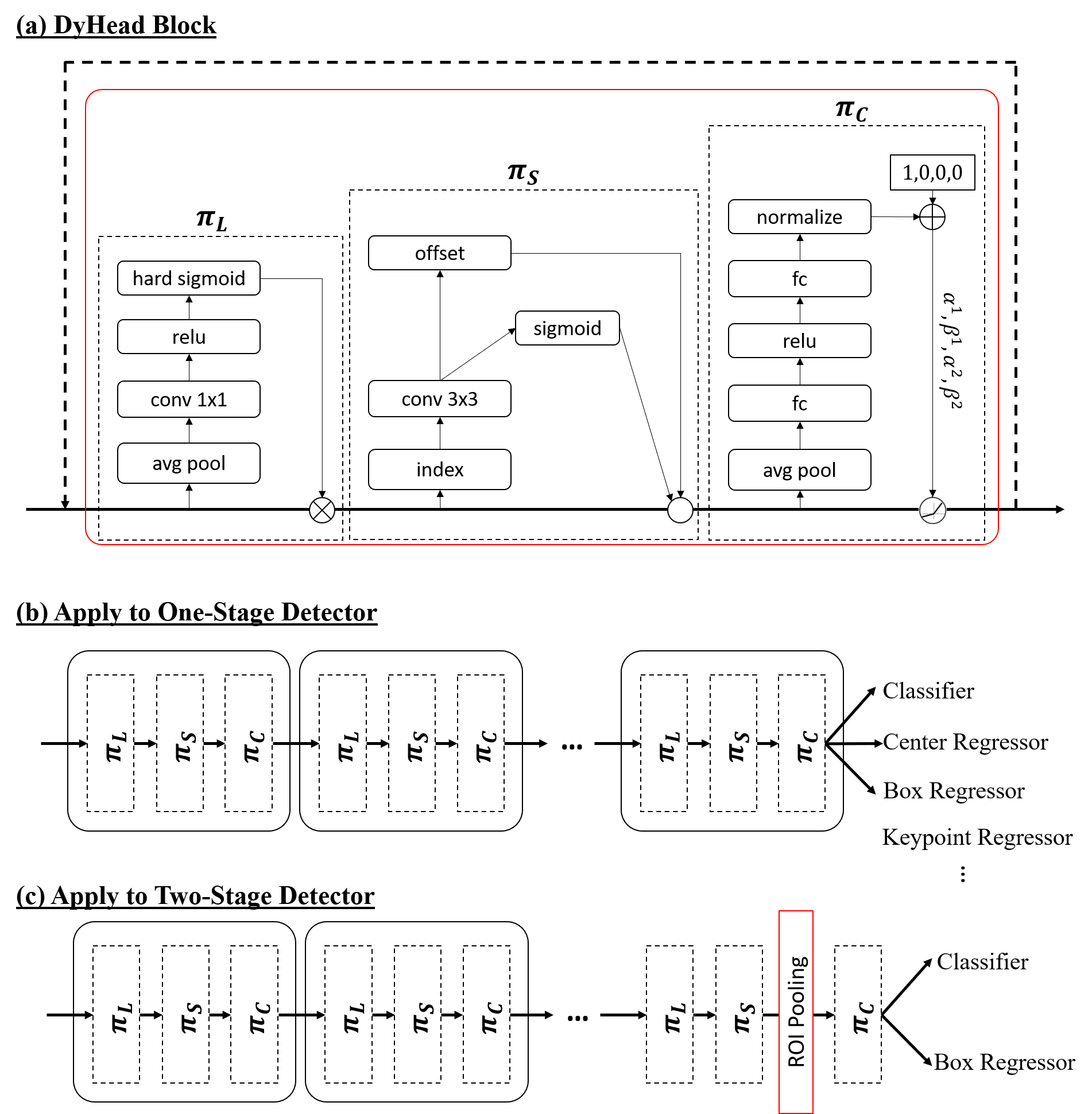}}
	\end{center}
	\caption{A detailed design of Dynamic Head. (a) shows the detailed implementation of each attention module. (b) shows how to apply our dynamic head blocks to one-stage object detector. (c) shows how to apply our dynamic head blocks to two-stage object detector.}
	\label{fig:dyhead}
\end{figure}

Finally, since the above three attention mechanisms are applied sequentially, we can nest Equation \ref{eq:main} multiple times to effectively stack multiple $\pi_{L}$, $\pi_{S}$, and $\pi_{C}$ blocks together. The detailed configuration of our dynamic head (\ie, \textit{DyHead} for simplification) block is shown in Figure \ref{fig:dyhead} (a).

As a summary, the whole paradigm of object detection with our proposed dynamic head is illustrated in Figure \ref{fig:main}. Any kinds of backbone network can be used to extract feature pyramid, which is further resized to the same scale, forming a 3-dimensional tensor $\mathcal{F}\in \mathcal{R}^{L\times S\times C}$, and then used as the input to the dynamic head. Next, several DyHead blocks including scale-aware, spatial-aware, and task-aware attentions are stacked sequentially. The output of the dynamic head can be used for different tasks and representations of object detection, such as classification, center/box regression, etc.. 

At the bottom of Figure \ref{fig:main}, we show the output of each type of attention. As we can see, the initial feature maps from backbones are noisy due to the domain difference from ImageNet pre-training. After passing through our scale-aware attention module, the feature maps become more sensitive to the scale differences of foreground objects; After further passing through our spatial-aware attention module, the feature maps become more sparse and focused on discriminative spatial locations of foreground objects. Finally, after passing through our task-aware attention module, the feature maps re-form into different activations based on the requirements of different down-stream tasks. 
These visualizations well demonstrate the effectiveness of each attention module.

\subsection{Generalizing to Existing Detectors}
In this section, we demonstrate how the proposed dynamic head can be integrated into existing detectors to effectively improve their performances. 

\paragraph{One-stage Detector.} One-stage detector predicts object locations by densely sampling locations from feature map, which simplifies the detector design. Typical one-stage detector (\eg, RetinaNet \cite{retinanet}) is composed of a backbone network to extract dense features and multiple task-specific sub-network branches to handle different tasks separately. As shown in previous work \cite{dyrelu}, object classification sub-network behaves very differently from bounding box regression sub-network. Controversial to this conventional approach, we only attach one unified branch instead of multiple branches to the backbone. It can handle multiple tasks simultaneously, thanks to the advantage of our multiple attention mechanisms. In this way, the architecture can be further simplified and the efficiency is improved as well. Recently, anchor-free variants of one-stage detectors became popular, for example, FCOS \cite{fcos}, ATSS \cite{atss} and RepPoint \cite{reppoints} re-formulated objects as centers and key-points to improve performance. Compared to RetinaNet, these methods require to attach a centerness prediction, or a keypoint prediction to either the classification branch or the regression branch, which makes the constructions of task-specific branches non-trivial. By contrast, deploying our dynamic head is more flexible since it only appends various types of predictions to the end of head, shown in Figure \ref{fig:dyhead} (b). 

\paragraph{Two-stage Detector.} Two-stage detectors utilize region proposal and ROI-pooling \cite{fasterrcnn} layers to extract intermediate representations from feature pyramid of a backbone network. To cooperate this characteristic, we first apply our scale-aware attention and spatial-aware attention on feature pyramid before a ROI-pooling layer and then use our task-aware attention to replace the original fully connected layers, as shown in Figure \ref{fig:dyhead} (c).

\subsection{Relation to Other Attention Mechanisms}

\paragraph{Deformable.} Deformable convolution \cite{deform, deformv2} has significantly improved the transformation learning of traditional convolutional layers by introducing sparse sampling. It has been widely used in object detection backbones to enhance the feature representations. Although it is rarely utilized in object detection head, we can regard it as solely modeling the $S$ sub-dimension in our representation. We find the deformable module used in the backbone can be complementary to the proposed dynamic head. In fact, with the deformable variant of ResNext-101-64x4d backbone, our dynamic head achieves a new state-of-the-art object detection result.

\paragraph{Non-local.} Non-Local Networks \cite{nl} is a pioneer work of utilizing attention modules to enhance the performance of object detection. However, it uses a simple formulation of dot-product to enhance a pixel feature by fusing other pixels’ features from different spatial locations. This behavior can be regarded as modeling only the $L \times S$ sub-dimensions in our representation.

\paragraph{Transformer.} Recently, there is a trend to introduce the Transformer module \cite{transformer} from natural language processing into computer vision tasks. Preliminary works \cite{detr, deform-detr, relationnet++} have demonstrated promising results in improving object detection. Transformer provides a simple solution to learn cross-attention correspondence and fuse features from different modalities by applying multi-head fully connected layers. This behavior can be viewed as modeling only the $S \times C$ sub-dimensions in our representation.

\paragraph{} The aforementioned three types of attention works only partially model sub-dimensions in the feature tensor. As a unified design, our dynamic head combines attentions on different dimensions into one coherent and efficient implementation. The following experiments show such a dedicated design can help existing object detectors achieve remarkable gains. Besides, our attention mechanisms explicitly address the challenges of object detection, in contrast to implicit working principles in existing solutions. 


\section{Experiment}
We evaluate our approach on the MS-COCO dataset \cite{coco} following the commonly used settings. MS-COCO contains 80 categories of around 160K images collected from the web. The dataset is split into the train2017, val2017, and test2017 subsets with 118K, 5K, 41K images respectively. The standard mean average precision ($AP$) metric is used to report results under different $IoU$ thresholds and object scales. In all our experiments, we only train on the train2017 images without using any extra data. For experiments of ablation studies, we evaluate the performances on the val2017 subset. When comparing to state-of-the-art methods, we report the official result returned from the test server on test-dev subset. 

\subsection{Implementation Details}
We implement our dynamic head block as a plugin, based on the popular implementation of Mask R-CNN benchmark \cite{maskrcnn}. If it is not specifically mentioned, our dynamic head is trained with the ATSS framework~\cite{atss} .  All models are trained using one compute node of 8 V100 GPUs each with 32GB memory.  

\paragraph{Training.} We use ResNet-50 as the model backbone in all ablation studies and train it with the standard 1x configuration. Other models are trained with the standard 2x training configurations as introduced in \cite{maskrcnn}. We use an initial learning rate of $0.02$ with weight decay of $1e-4$ and momentum of $0.9$ . The learning rate is stepped down by a factor of $0.1$ at the $67\%$ and $89\%$ of training epochs. Standard augmentation with random horizontal flipping is used. To compare with previous methods trained with multi-scale inputs, we also conduct multi-scale training for selective models.

\paragraph{Inference.} To compare with state-of-the-art methods reported using test time augmentation, we also evaluate our best model with multi-scale testing. Other tricks, such as model EMA, mosaic, mix-up, label smoothing, soft-NMS or adaptive multi-scale testing \cite{sniper}, are not used.

\subsection{Ablation Study}
We conduct a series of ablation studies to demonstrate the effectiveness and efficiency of our dynamic head.

\begin{table}[h]
\centering
\setlength{\tabcolsep}{6pt}
        \small
            \renewcommand{\arraystretch}{1.3}
\begin{tabular}{|ccc|ccc|ccc|}
\hline
 L. & S. & C. & AP & AP$_{50}$ & AP$_{75}$ & AP$_{S}$ & AP$_{M}$ & AP$_{L}$ \\
\hline\hline
$\times$ & $\times$ & $\times$  & 39.0 & 57.2 & 42.4 & 22.1 & 43.1 & 50.2\\
\hline
$\checkmark$ & $\times$ & $\times$  & 39.9 & 57.8 & 43.5 & 25.4 & 44.0 & 52.4\\
$\times$ & $\checkmark$ & $\times$  & 41.4 & 58.5 & 45.2 & 26.8 & 45.2 & 54.3\\
$\times$ & $\times$ & $\checkmark$  & 40.3 & 58.3 & 43.9 & 24.2 & 44.6 & 53.7\\
\hline
$\times$ & $\checkmark$ & $\checkmark$ &  42.0 & 59.5 & 45.5 & 25.5 & 46.1 & 55.2\\
$\checkmark$ & $\times$ & $\checkmark$ &  40.6 & 58.6 & 44.4 & 24.6 & 44.8 & 53.3\\
$\checkmark$ & $\checkmark$ & $\times$ &  41.9 & 59.2 &  45.6 & 24.8 & 46.1 & 54.5\\
\hline
$\checkmark$ & $\checkmark$ & $\checkmark$ & \bf{42.6} & \bf{60.1} & \bf{46.4} & \bf{26.1} & \bf{46.8} & \bf{56.0}\\
\hline
\end{tabular}
\vspace{0.8em}
\caption{Ablation study on the effectiveness of each attention module in our dynamic head block.}
\vspace{-0.8em}
\label{tb:ab1}
\end{table}

\paragraph{Effectiveness of Attention Modules.}
We first conduct a controlled study on the effectiveness of different components in our dynamic head block by gradually adding them to the baseline. As shown in Table \ref{tb:ab1}, ``L.", ``S.", ``C." represent our scale-aware attention module, spatial-aware attention module, and task-aware module, respectively. We can observe that individually adding each component to the baseline implementation improves its performance by $0.9$ $AP$, $2.4$ $AP$ and $1.3$ $AP$. It is expected to see the spatial-aware attention module archives the biggest gain because of its dominant dimensionality among three modules. When we add both ``L." and ``S" to the baseline, it continuously improves the performance by $2.9$ $AP$. Finally, our full dynamic head block significantly improves the baseline by $3.6$ $AP$. This experiment demonstrates that different components work as a coherent module.

\begin{figure}[t]
	\begin{center}
		\hbox{\includegraphics[width=0.95\linewidth]{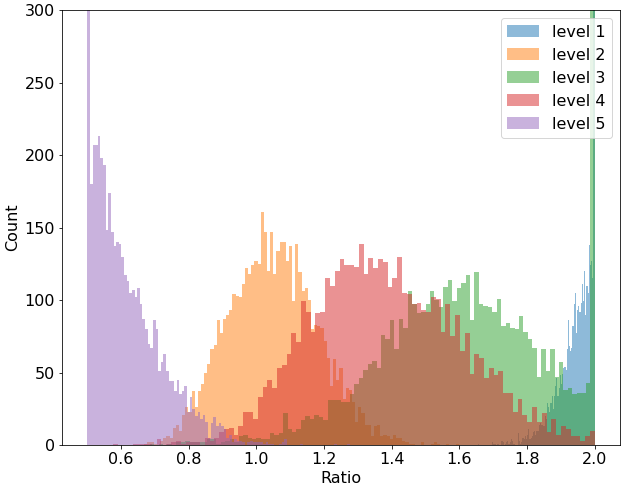}}
	\end{center}\vspace{-1.5em}
	\caption{Ablation study on the effectiveness of our scale-aware attention module.}
	\label{fig:level}
\end{figure}

\paragraph{Effectiveness on Attention Learning.} We then demonstrate the effectiveness of attention learning in our dynamic head module. Figure \ref{fig:level} shows the trend of the learned scale ratios (calculated by dividing the learned weight of higher resolution by the learned weight of lower resolution) on different level of features in our scale-aware attention module. The histogram is calculated using all images from the COCO val2017 subset. It is clear to see that our scale-aware attention module tends to regulate higher resolution feature maps ("level 5" purple histogram in the figure) toward lower resolution and lower resolution feature maps ("level 1" blue histogram in the figure) toward higher resolution to smooth the scale discrepancy form different feature levels. This proves the effectiveness of scale-aware attention learning.

\begin{figure}[t]
	\begin{center}
		\includegraphics[width=1.0\linewidth]{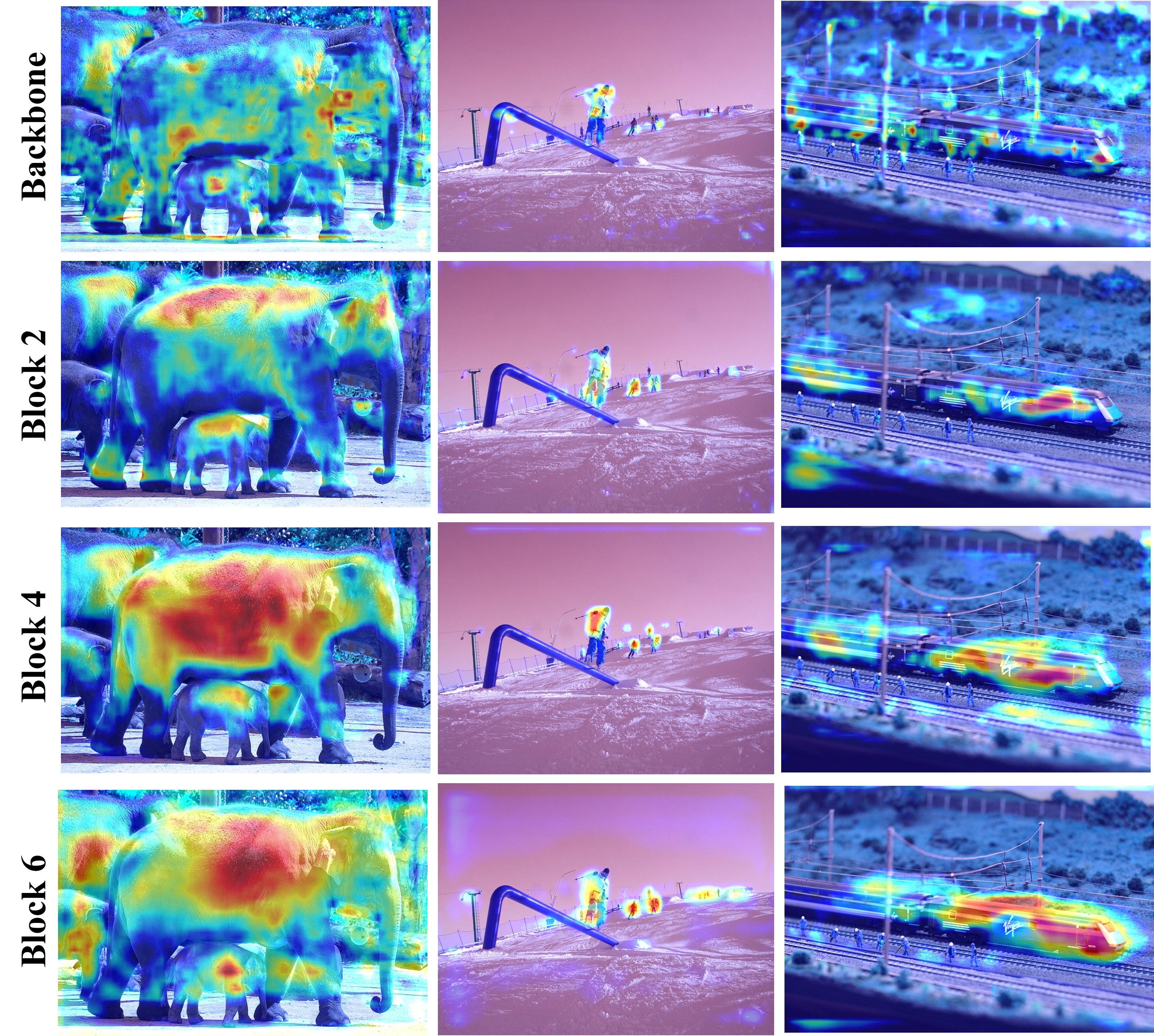}
	\end{center} 
	\caption{A visualization on the effectiveness of our spatial-aware attention module.}
	\label{fig:viz}
\end{figure}

Figure \ref{fig:viz} visualizes the feature map output before and after applying different number (i.e. 2,4,6) of blocks of attention modules.  Before applying our attention modules, the feature maps extracted from the backbone are very noisy and fail to focus on the foreground objects. As the feature maps pass through more attention modules (from block 2 to block 6 as shown in the figure), it is obvious to see the feature maps cover more foreground objects and focus more accurately on their discriminative spatial locations. This visualization well demonstrates the effectiveness of the spatial-aware attention learning

\paragraph{Efficiency on the Depth of Head.} We evaluate the efficiency of our dynamic head by controlling the depth (number of blocks). As shown in Table \ref{tb:ab2}, we vary the number of used DyHead blocks (\eg, $1$, $2$, $4$, $8$, $10$ blocks) and compare their performances and computational costs (GFLOPs) with the baseline. Our dynamic head can benefit from the increase of depth by stacking more blocks until $8$. It is worth noting that our method with $2$ blocks has already outperformed the baseline at even lower computation cost. Meanwhile, even with $6$ blocks, the increment of computational cost is negligible compared to the computation cost of the backbone, while largely improving the accuracy. It demonstrates the efficiency of our method.

\begin{table}[t]  
\centering
\setlength{\tabcolsep}{9pt}
        \small
            \renewcommand{\arraystretch}{1.2}
\begin{tabular}{|c|c|ccc|}
\hline
~ \#Block ~ & ~ GFLOPs~ & ~ AP ~ & ~ AP$_{50}$ ~ & AP$_{75}$ \\
\hline\hline
Baseline & 254.39  & 39.0 & 57.2 & 42.4 \\
\hline
1 & -84.69  & 36.7 & 55.5 & 40.0 \\
2 & -63.45  & 39.5 & 57.8 & 43.1 \\
4 & -20.97  & 42.0 & 59.9 & 45.9 \\
\bf{6} & +21.50  & \bf{42.6} & \bf{60.1} & \bf{46.4} \\
8 & +63.98  & 42.5 & 59.6 & 46.1 \\
10 & +106.46 & 42.3 & 59.4 & 45.9 \\
\hline
\end{tabular} \vspace{1.0em}
\caption{Ablation study on the efficiency and effectiveness of stacking different number of dynamic head blocks.}
\label{tb:ab2}
\end{table}

\paragraph{Generalization on Existing Object Detectors.} We evaluate the generalization ability of the dynamic head by plugging it to popular object detectors, such as Faster-RCNN~\cite{fasterrcnn}, RetinaNet~\cite{retinanet}, ATSS~\cite{atss}, FCOS~\cite{fcos}, and RepPoints~\cite{reppoints}. These methods represent a wide variety of object detection frameworks (\eg, two-stage vs. one-stage, anchor-based vs. anchor-free, box-based vs. point-based). As shown in Table \ref{tb:ab3}, our dynamic head significantly boosts all popular object detectors by $1.2$ $\sim$ $3.2$ $AP$. It demonstrates the generality of our method.

\begin{table}[t]  
\centering
\setlength{\tabcolsep}{12pt}
        \small
            \renewcommand{\arraystretch}{1.3}
\begin{tabular}{|c|ccc|}
\hline
~ Method ~  & ~ AP ~ & ~ AP$_{50}$ ~ & AP$_{75}$\\
\hline\hline
\multicolumn{4}{|l|}{\it{anchor-based two-stage:}}\\
\hline
Faster R-CNN~\cite{fasterrcnn}   & 36.4 & 57.9 & 39.4\\
\quad + DyHead   & \bf{38.9} & \bf{57.6} & \bf{42.0} \\
\hline\hline
\multicolumn{4}{|l|}{\it{anchor-based one-stage:}}\\
\hline
RetinaNet~\cite{retinanet}  & 35.7 & 54.3 & 37.9\\
\quad + DyHead  & \bf{38.4} & \bf{57.5} & \bf{41.3}  \\
\hline\hline
\multicolumn{4}{|l|}{\it{anchor-free box-based:}}\\
\hline
ATSS~\cite{atss} &  39.4 & 57.5 & 42.9 \\
\quad + DyHead  & \bf{42.6} & \bf{60.1} & \bf{46.4} \\
\hline\hline
\multicolumn{4}{|l|}{\it{anchor-free center-based:}}\\
\hline
FCOS~\cite{fcos} &  38.8 & 57.3 & 41.9 \\
\quad + DyHead  & \bf{40.0} & \bf{58.2} & \bf{43.4}  \\
\hline\hline
\multicolumn{4}{|l|}{\it{anchor-free keypoint-based:}}\\
\hline
RepPoints~\cite{reppoints}   & 38.2 & 59.7 & 40.7\\
\quad + DyHead   & \bf{39.6} & \bf{59.8} & \bf{42.8}\\
\hline
\end{tabular}\vspace{0.8em}
\caption{Ablation study on the generalization of our  dynamic head when applying to popular object detection methods.}
\label{tb:ab3}
\end{table} 

\subsection{Comparison with the State of the Art}

We compare the performance of the dynamic head with several standard backbones and state-of-the-art object detectors. 

\paragraph{Cooperate with Different Backbones.} We first demonstrate the compatibility of dynamic head with different backbones. As shown in Table \ref{tb:dets}, we evaluate the performances of object detector by integrating dynamic head with the ResNet-50, ResNet-101 and ResNeXt-101 backbones, and compare with recent methods with similar configurations, including Mask R-CNN~\cite{maskrcnn}, Cascade-RCNN~\cite{cascade}, FCOS~\cite{fcos}, ATSS~\cite{atss} and BorderDet~\cite{borderdet}. Our method consistently outperforms previous methods with a big margin. When compared to the best detector BorderDet~\cite{borderdet} with same settings, our method outperforms it by $1.1$ $AP$ with the ResNet-101 backbone and by $1.2$ $AP$ with the ResNeXt-64x4d-101 backbone, where the improvement is significant due to the challenges in the COCO benchmark. 

\begin{table*}[h]  
\centering
\setlength{\tabcolsep}{9.5pt}
        \small
            \renewcommand{\arraystretch}{1.2}
\begin{tabular}{|c|c|c|ccc|ccc|}
\hline
 ~ Method ~ &  ~ Backbone ~  &  ~ Iteration ~ & ~ AP ~ & ~ AP$_{50}$ ~ & ~ AP$_{75}$ ~ & ~ AP$_{S}$ ~ & AP$_{M}$ ~ & ~ AP$_{L}$ ~ \\
\hline\hline
\multicolumn{9}{|l|}{\it{two-stage detector:}}\\
\hline
Mask R-CNN\cite{maskrcnn} & ResNet-101 & 2x & 38.2 & 60.3 & 41.7 & 20.1 & 41.1 & 50.2 \\
Cascade-RCNN\cite{cascade} & ResNet-50 & 3x & 40.6 & 59.9 & 44.0 & 22.6 & 42.7 & 52.1 \\
Cascade-RCNN\cite{cascade} & ResNet-101 & 3x & 42.8 & 62.1 & 46.3 & 23.7 & 45.5 & 55.2 \\
\hline\hline
\multicolumn{9}{|l|}{\it{one-stage detector:}}\\
\hline
FCOS\cite{fcos} & ResNet-101 & 2x & 41.5 & 60.7 & 45.0 & 24.4 & 44.8 & 51.6 \\
FCOS\cite{fcos} & ResNeXt-64x4d-101 & 2x & 43.2 & 62.8 & 46.6 & 26.5 & 46.2 & 53.3 \\
ATSS\cite{atss} & ResNet-101 & 2x & 43.6 & 62.1 & 47.4 & 26.1 & 47.0 & 53.6 \\
ATSS\cite{atss} & ResNeXt-64x4d-101 & 2x & 45.6 & 64.6 & 49.7 & 28.5 & 48.9 & 55.6 \\
BorderDet\cite{borderdet} & ResNet-101 & 1x & 43.2 & 62.1 & 46.7 & 24.4 & 46.3 & 54.9 \\
BorderDet\cite{borderdet} & ResNet-101 & 2x & 45.4 & 64.1 & 48.8 & 26.7 & 48.3 & 56.5 \\
BorderDet\cite{borderdet} & ResNeXt-64x4d-101 & 2x & 46.5 & 65.7 & 50.5 & 29.1 & 49.4 & 57.5 \\
\hline
\bf{DyHead} & ResNet-50 & 1x & 43.0 & 60.7 & 46.8 & 24.7 & 46.4 & 53.9 \\
\bf{DyHead} & ResNet-101 & 2x & 46.5 & 64.5 & 50.7 & 28.3& 50.3 & 57.5 \\
\bf{DyHead} & ResNeXt-64x4d-101 & 2x & 47.7 & 65.7 & 51.9 & 31.5 & 51.7 & 60.7 \\
\hline
\end{tabular} \vspace{0.3em}
\caption{Comparison with results using different backbones on the MS COCO test-dev set} \vspace{0.6em}
\label{tb:dets}
\end{table*} 

\begin{table*}[h]  
\centering
\setlength{\tabcolsep}{7.5pt}
        \small
            \renewcommand{\arraystretch}{1.2}
\begin{tabular}{|c|c|c|ccc|ccc|}
\hline
 ~ Method ~ &  ~ Backbone ~  &  ~ Iteration ~ & ~ AP ~ & ~ AP$_{50}$ ~ & ~ AP$_{75}$ ~ & ~ AP$_{S}$ ~ & AP$_{M}$ ~ & ~ AP$_{L}$ ~ \\
\hline\hline
\multicolumn{9}{|l|}{\it{multi-scale training:}}\\
\hline
ATSS\cite{atss} & ResNeXt-64x4d-101-DCN & 2x & 47.7 & 66.5 & 51.9 & 29.7 & 50.8 & 59.4 \\
SEPC\cite{spec} & ResNeXt-64x4d-101-DCN & 2x & 50.1 & 69.8 & 54.3 & 31.3 & 53.3 & 63.7 \\
BorderDet\cite{borderdet} & ResNeXt-64x4d-101-DCN & 2x & 48.0 & 67.1 & 52.1 & 29.4 & 50.7 & 60.5 \\
RepPoints v2\cite{reppointsv2} & ResNeXt-64x4d-101-DCN & 2x & 49.4 & 68.9 & 53.4 & 30.3 & 52.1 & 62.3 \\
RelationNet++\cite{relationnet++} & ResNeXt-64x4d-101-DCN & 2x & 50.3 & 69.0 & 55.0 & 32.8 & 55.0 & 65.8 \\
DETR\cite{detr} & ResNet-101 & $\sim$25x & 44.9 & 64.7 & 47.7 & 23.7 & 49.5 & 62.3 \\
Deformable DETR\cite{deform-detr} & ResNeXt-64x4d-101-DCN & $\sim$4x & 50.1 & 69.7 & 54.6 & 30.6 & 52.8 & 64.7 \\
EfficientDet\cite{efficientdet} & Efficient-B7 & $\sim$50x & 52.2 & 71.4 & 56.3 & -- & -- & -- \\
SpineNet\cite{spinenet} & SpineNet-190 & $\sim$40x & 52.1 & 71.8 & 56.5 & 35.4 & 55.0 & 63.6 \\
\hline
\bf{DyHead} & ResNeXt-64x4d-101-DCN & 2x & \bf{52.3} & \bf{70.7} & \bf{57.2} & \bf{35.1} & \bf{56.2} & \bf{63.4}\\
\hline\hline
\multicolumn{9}{|l|}{\it{multi-scale training and multi-scale testing:}}\\
\hline
ATSS\cite{atss} & ResNeXt-64x4d-101-DCN & 2x & 50.7 & 68.9 & 56.3 & 33.2 & 52.9 & 62.4 \\
BorderDet\cite{borderdet} & ResNeXt-64x4d-101-DCN & 2x & 50.3 & 68.9 & 55.2 & 32.8 & 52.8 & 62.3 \\
RepPoints v2\cite{reppointsv2} & ResNeXt-64x4d-101-DCN & 2x & 52.1 & 70.1 & 57.5 & 34.5 & 54.6 & 63.6 \\
Deformable DETR\cite{deform-detr} & ResNeXt-64x4d-101-DCN & $\sim$4x & 52.3 & 71.9 & 58.1 & 34.4 & 54.4 & 65.6 \\
RelationNet++\cite{relationnet++} & ResNeXt-64x4d-101-DCN & 2x & 52.7 & 70.4 & 58.3 & 35.8 & 55.3 & 64.7 \\
\hline
\bf{DyHead} & ResNeXt-64x4d-101-DCN & 2x & \bf{54.0} & \bf{72.1} & \bf{59.3} & \bf{37.1} & \bf{57.2} &  \bf{66.3}\\
\hline
\end{tabular}\vspace{0.8em}
\caption{Comparison with the state-of-the-art results on the MS COCO test-dev set}
\label{tb:stoa}
\end{table*}

\paragraph{Compared to State-of-the-Art Detectors.} We compare our methods with state-of-the-art detectors~\cite{atss, spec, borderdet, reppointsv2, detr, efficientdet, spinenet}, including some concurrent works \cite{deform-detr, relationnet++}.
As shown in Table \ref{tb:stoa}, we summarize these existing work into two categories: one using multi-scale training, and the other using both multi-scale training and multi-scale testing. 

Compared with methods with only multi-scale training, our method achieves a new state of the art at $52.3$ $AP$ with only 2x training schedule. Our method is competitive and more efficient to learn compared with EffcientDet \cite{efficientdet} and SpineNet \cite{spinenet}, with a significantly less $1/20$ training time. Compared with the latest work \cite{detr, deform-detr, relationnet++}, which utilize Transformer modules as attention, our dynamic head is superior to these methods with more than 2.0 $AP$ gain, while using less training time than theirs. It demonstrates that our dynamic head can coherently combine multiple modalities of attentions from different perspectives into a unified head, resulting in better efficiency and effectiveness.

We further compare our method with state-of-the-art results \cite{atss, borderdet, reppointsv2, deform-detr, relationnet++} with test time augmentation (TTA) using both multi-scale training and multi-scale testing. Our dynamic head helps achieve a new state-of-the-art result at $54.0$ $AP$, which significantly outperforms concurrent best methods by $1.3$ $AP$.

\section{Conclusion}

In this paper, we have presented a novel object detection head, which unify the scale-aware, spatial-aware, and task-aware attentions in a single framework. It suggests a new view of object detection head with attentions. As a plugin block, the dynamic head can be flexibly integrated into any existing object detector framework to boost its performance. Moreover, it is efficient to learn. Our study shows that designing and learning attentions in the object detection head is an interesting direction which deserves more focused studies. This work only takes a step, and could be further improved in these aspects: how to make the full attention model easy to learn and efficient to compute, and how to systematically consider more modalities of attentions into the head designing for better performance.

\begin{table*}[t]  
\centering
\setlength{\tabcolsep}{9.5pt}
        \small
            \renewcommand{\arraystretch}{1.2}
\begin{tabular}{|c|c|c|ccc|ccc|}
\hline
 ~ Method ~ &  ~ Backbone ~ & Iteration &  ~ AP &  ~ AP$_{50}$  &  ~ AP$_{75}$ &  ~ AP$_{S}$  & ~ AP$_{M}$  &  ~ AP$_{L}$ \\
\hline\hline
Mask R-CNN\cite{maskrcnn} & Swin-T & 3x & 46.0 & 68.1 & 50.3 & 31.2 & 49.2 & 60.1 \\
Cascade Mask R-CNN\cite{cascade} & Swin-T & 3x & 50.4 & 69.2 & 54.7 & 33.8 & 54.1 & 65.2 \\
RepPoints v2\cite{reppointsv2} & Swin-T & 3x & 50.0 & 68.5 & 54.2 & -- & -- & -- \\
SparseRCNN\cite{sparsercnn} & Swin-T & 3x & 47.9 & 67.3 & 52.3 & -- & -- & -- \\
ATSS\cite{atss} & Swin-T & 3x & 47.2 & 66.5 & 51.3 & -- & -- & -- \\
\hline
\bf{DyHead} & Swin-T & 2x & 49.7 & 68.0 & 54.3 & 33.3 & 54.2 & 64.2 \\
\hline
\end{tabular} \vspace{0.3em}
\caption{Comparison with results using transformer backbone on the MS COCO validation set.} \vspace{0.6em}
\label{tb:trans1}
\end{table*}

\begin{table*}[t]  
\centering
\setlength{\tabcolsep}{9.5pt}
        \small
            \renewcommand{\arraystretch}{1.2}
\begin{tabular}{|c|c|c|c|ccc|ccc|}
\hline
 ~ Method ~ &  ~ Backbone ~  & Iteration & AP$_{val}$ &  AP & AP$_{50}$ & AP$_{75}$ & AP$_{S}$ & AP$_{M}$ & AP$_{L}$ \\
\hline\hline
CenterNet2$\dagger$ \cite{centernet2} & Res2Net-101-DCN & 8x & 56.1 & 56.4 & 74.0 & 61.6 & 38.7 & 59.7 & 68.6 \\
CopyPaste$\dagger$ \cite{copypaste} & Efficient-B7 & 8x & 57.0 & 57.3 & -- & -- & -- & -- & -- \\
HTC++\cite{swin} & Swin-L & 6x & 58.0 & 58.7 & -- & -- & -- & -- & -- \\
\hline
\bf{DyHead} & Swin-L & 2x & 58.4 & 58.7 & 77.1 & 64.5 & 41.7 & 62.0 & 72.8 \\
\bf{DyHead$\dagger$} & Swin-L & 2x & \bf{60.3} & \bf{60.6} & \bf{78.5} & \bf{66.6} & \bf{43.9} & \bf{64.0} & \bf{74.2} \\
\hline
\end{tabular} \vspace{0.3em}
\caption{Comparison with latest methods on the MS COCO test-dev set. $\dagger$ demonstrates method with extra data.} \vspace{0.6em}
\label{tb:latest}
\end{table*} 

\section*{Appendix}

We keep improving our performance after submission. Recently, there is a hot trend on adapting transformers as vision backbones and demonstrating promising performance. When training our dynamic head with latest transformer backbone \cite{swin}, extra data and increased input size, we can further improve the current SOTA on COCO benchmark.

\paragraph{Cooperate with Transformer Backbones.}
We cooperate our dynamic head with the latest transformer-based backbones, such as \cite{swin}. Shown in Table \ref{tb:trans1}, our dynamic head is competitive to \cite{cascade} which requires extra mask ground-truth to help boost performance. Meanwhile, compared to the baseline method \cite{atss} used in our framework, we further improve its performance by $2.5$ $AP$. This well proves that our dynamic head is complementary to transformer-based backbone to further improve its performance on downstream object detection task.  

\paragraph{Cooperate with Larger Inputs and Extra Data.} 
We find that our dynamic head can further benefit from larger input size and extra data generated using self-training method \cite{self-training}. We increase the maximum image side from 1333 to 2000 and use a multi-scale training with minimum image side varying from 480 to 1200. Similar to the training scheme described in section 4.1, we avoid using more tricks to ensure reproducibility. As shown in Table \ref{tb:latest}, our dynamic head leads significant gain compared to latest works \cite{copypaste, centernet2} and matches the performance of \cite{swin} without using extra mask ground-truth. Meanwhile, our dynamic head requires less than $1/3$ of training time of these works. This demonstrates our superior efficiency and effectiveness. Furthermore, we follow \cite{self-training} to generate pseudo labels on ImageNet dataest and use it as an extra data. Our dynamic head can largely benefit from large scale data and further improve the COCO state-of-the-art result to a new record high at $60.6$ $AP$.

{\small
\bibliographystyle{ieee_fullname}
\bibliography{egbib}

\begin{thebibliography}{10}\itemsep=-1pt

\bibitem{cascade}
Zhaowei Cai and N. Vasconcelos.
\newblock Cascade r-cnn: Delving into high quality object detection.
\newblock {\em 2018 IEEE/CVF Conference on Computer Vision and Pattern
  Recognition}, pages 6154--6162, 2018.

\bibitem{detr}
Nicolas Carion, F. Massa, Gabriel Synnaeve, Nicolas Usunier, Alexander
  Kirillov, and Sergey Zagoruyko.
\newblock End-to-end object detection with transformers.
\newblock {\em ArXiv}, abs/2005.12872, 2020.

\bibitem{dyrelu}
Y. Chen, X. Dai, Mengchen Liu, Dongdong Chen, Lu Yuan, and Zicheng Liu.
\newblock Dynamic relu.
\newblock {\em ArXiv}, abs/2003.10027, 2020.

\bibitem{reppointsv2}
Y. Chen, Zheng Zhang, Yue Cao, L. Wang, Stephen Lin, and H. Hu.
\newblock Reppoints v2: Verification meets regression for object detection.
\newblock {\em ArXiv}, abs/2007.08508, 2020.

\bibitem{relationnet++}
Cheng Chi, Fangyun Wei, and Han Hu.
\newblock Relationnet++: Bridging visual representations for object detection
  via transformer decoder.
\newblock {\em ArXiv}, abs/2010.15831, 2020.

\bibitem{rcnn}
Jifeng Dai, Y. Li, Kaiming He, and Jian Sun.
\newblock R-fcn: Object detection via region-based fully convolutional
  networks.
\newblock {\em ArXiv}, abs/1605.06409, 2016.

\bibitem{deform}
Jifeng Dai, Haozhi Qi, Y. Xiong, Y. Li, Guodong Zhang, H. Hu, and Y. Wei.
\newblock Deformable convolutional networks.
\newblock {\em 2017 IEEE International Conference on Computer Vision (ICCV)},
  pages 764--773, 2017.

\bibitem{spinenet}
Xianzhi Du, Tsung-Yi Lin, Pengchong Jin, Golnaz Ghiasi, Mingxing Tan, Yin Cui,
  Quoc~V. Le, and Xiaodan Song.
\newblock Spinenet: Learning scale-permuted backbone for recognition and
  localization, 2020.

\bibitem{centernet}
Kaiwen Duan, S. Bai, Lingxi Xie, H. Qi, Q. Huang, and Q. Tian.
\newblock Centernet: Keypoint triplets for object detection.
\newblock {\em 2019 IEEE/CVF International Conference on Computer Vision
  (ICCV)}, pages 6568--6577, 2019.

\bibitem{copypaste}
Golnaz Ghiasi, Yin Cui, Aravind Srinivas, Rui Qian, Tsung-Yi Lin, Ekin~D.
  Cubuk, Quoc~V. Le, and Barret Zoph.
\newblock Simple copy-paste is a strong data augmentation method for instance
  segmentation, 2020.

\bibitem{fastrcnn}
Ross~B. Girshick.
\newblock Fast r-cnn.
\newblock {\em 2015 IEEE International Conference on Computer Vision (ICCV)},
  pages 1440--1448, 2015.

\bibitem{maskrcnn}
Kaiming He, Georgia Gkioxari, Piotr Doll{\'a}r, and Ross~B. Girshick.
\newblock Mask r-cnn.
\newblock {\em 2017 IEEE International Conference on Computer Vision (ICCV)},
  pages 2980--2988, 2017.

\bibitem{resnet}
Kaiming He, X. Zhang, Shaoqing Ren, and Jian Sun.
\newblock Deep residual learning for image recognition.
\newblock {\em 2016 IEEE Conference on Computer Vision and Pattern Recognition
  (CVPR)}, pages 770--778, 2016.

\bibitem{aug}
A. Krizhevsky, Ilya Sutskever, and Geoffrey~E. Hinton.
\newblock Imagenet classification with deep convolutional neural networks.
\newblock In {\em CACM}, 2017.

\bibitem{fpn}
Tsung-Yi Lin, Piotr Doll{\'a}r, Ross~B. Girshick, Kaiming He, Bharath
  Hariharan, and Serge~J. Belongie.
\newblock Feature pyramid networks for object detection.
\newblock {\em 2017 IEEE Conference on Computer Vision and Pattern Recognition
  (CVPR)}, pages 936--944, 2017.

\bibitem{retinanet}
Tsung-Yi Lin, Priyal Goyal, Ross~B. Girshick, Kaiming He, and Piotr Doll{\'a}r.
\newblock Focal loss for dense object detection.
\newblock {\em IEEE Transactions on Pattern Analysis and Machine Intelligence},
  42:318--327, 2020.

\bibitem{coco}
Tsung-Yi Lin, Michael Maire, Serge Belongie, Lubomir Bourdev, Ross Girshick,
  James Hays, Pietro Perona, Deva Ramanan, C.~Lawrence Zitnick, and Piotr
  Dollár.
\newblock Microsoft coco: Common objects in context, 2014.
\newblock cite arxiv:1405.0312Comment: 1) updated annotation pipeline
  description and figures; 2) added new section describing datasets splits; 3)
  updated author list.

\bibitem{panet}
Shu Liu, Lu Qi, Haifang Qin, J. Shi, and J. Jia.
\newblock Path aggregation network for instance segmentation.
\newblock {\em 2018 IEEE/CVF Conference on Computer Vision and Pattern
  Recognition}, pages 8759--8768, 2018.

\bibitem{swin}
Ze Liu, Yutong Lin, Yue Cao, Han Hu, Yixuan Wei, Zheng Zhang, Stephen Lin, and
  Baining Guo.
\newblock Swin transformer: Hierarchical vision transformer using shifted
  windows, 2021.

\bibitem{libra}
Jiangmiao Pang, K. Chen, J. Shi, H. Feng, Wanli Ouyang, and D. Lin.
\newblock Libra r-cnn: Towards balanced learning for object detection.
\newblock {\em 2019 IEEE/CVF Conference on Computer Vision and Pattern
  Recognition (CVPR)}, pages 821--830, 2019.

\bibitem{borderdet}
Han Qiu, Yuchen Ma, Zeming Li, Songtao Liu, and J. Sun.
\newblock Borderdet: Border feature for dense object detection.
\newblock In {\em ECCV}, 2020.

\bibitem{yolo}
Joseph Redmon, S. Divvala, Ross~B. Girshick, and Ali Farhadi.
\newblock You only look once: Unified, real-time object detection.
\newblock {\em 2016 IEEE Conference on Computer Vision and Pattern Recognition
  (CVPR)}, pages 779--788, 2016.

\bibitem{fasterrcnn}
Shaoqing Ren, Kaiming He, Ross~B. Girshick, and J. Sun.
\newblock Faster r-cnn: Towards real-time object detection with region proposal
  networks.
\newblock {\em IEEE Transactions on Pattern Analysis and Machine Intelligence},
  39:1137--1149, 2015.

\bibitem{snip}
B. Singh and L. Davis.
\newblock An analysis of scale invariance in object detection - snip.
\newblock {\em 2018 IEEE/CVF Conference on Computer Vision and Pattern
  Recognition}, pages 3578--3587, 2018.

\bibitem{sniper}
B. Singh, Mahyar Najibi, and L. Davis.
\newblock Sniper: Efficient multi-scale training.
\newblock In {\em NeurIPS}, 2018.

\bibitem{sparsercnn}
Peize Sun, Rufeng Zhang, Yi Jiang, Tao Kong, Chenfeng Xu, Wei Zhan, Masayoshi
  Tomizuka, Lei Li, Zehuan Yuan, Changhu Wang, and Ping Luo.
\newblock {SparseR-CNN}: End-to-end object detection with learnable proposals.
\newblock {\em arXiv preprint arXiv:2011.12450}, 2020.

\bibitem{efficientdet}
Mingxing Tan, R. Pang, and Quoc~V. Le.
\newblock Efficientdet: Scalable and efficient object detection.
\newblock {\em 2020 IEEE/CVF Conference on Computer Vision and Pattern
  Recognition (CVPR)}, pages 10778--10787, 2020.

\bibitem{fcos}
Zhi Tian, Chunhua Shen, Hao Chen, and Tong He.
\newblock Fcos: Fully convolutional one-stage object detection.
\newblock {\em 2019 IEEE/CVF International Conference on Computer Vision
  (ICCV)}, pages 9626--9635, 2019.

\bibitem{transformer}
Ashish Vaswani, Noam Shazeer, Niki Parmar, Jakob Uszkoreit, Llion Jones,
  Aidan~N. Gomez, L. Kaiser, and Illia Polosukhin.
\newblock Attention is all you need.
\newblock {\em ArXiv}, abs/1706.03762, 2017.

\bibitem{nl}
X. Wang, Ross~B. Girshick, A. Gupta, and Kaiming He.
\newblock Non-local neural networks.
\newblock {\em 2018 IEEE/CVF Conference on Computer Vision and Pattern
  Recognition}, pages 7794--7803, 2018.

\bibitem{spec}
Xinjiang Wang, S. Zhang, Zhuoran Yu, Litong Feng, and Wayne Zhang.
\newblock Scale-equalizing pyramid convolution for object detection.
\newblock {\em 2020 IEEE/CVF Conference on Computer Vision and Pattern
  Recognition (CVPR)}, pages 13356--13365, 2020.

\bibitem{resnext}
Saining Xie, Ross~B. Girshick, Piotr Doll{\'a}r, Zhuowen Tu, and Kaiming He.
\newblock Aggregated residual transformations for deep neural networks.
\newblock {\em 2017 IEEE Conference on Computer Vision and Pattern Recognition
  (CVPR)}, pages 5987--5995, 2017.

\bibitem{reppoints}
Ze Yang, S. Liu, H. Hu, Liwei Wang, and Stephen Lin.
\newblock Reppoints: Point set representation for object detection.
\newblock {\em 2019 IEEE/CVF International Conference on Computer Vision
  (ICCV)}, pages 9656--9665, 2019.

\bibitem{dilated}
F. Yu and V. Koltun.
\newblock Multi-scale context aggregation by dilated convolutions.
\newblock {\em CoRR}, abs/1511.07122, 2016.

\bibitem{atss}
Shifeng Zhang, Cheng Chi, Yongqiang Yao, Z. Lei, and S. Li.
\newblock Bridging the gap between anchor-based and anchor-free detection via
  adaptive training sample selection.
\newblock {\em 2020 IEEE/CVF Conference on Computer Vision and Pattern
  Recognition (CVPR)}, pages 9756--9765, 2020.

\bibitem{centernet2}
Xingyi Zhou, Vladlen Koltun, and Philipp Kr{\"a}henb{\"u}hl.
\newblock Probabilistic two-stage detection.
\newblock In {\em arXiv preprint arXiv:2103.07461}, 2021.

\bibitem{deformv2}
X. Zhu, H. Hu, Stephen Lin, and Jifeng Dai.
\newblock Deformable convnets v2: More deformable, better results.
\newblock {\em 2019 IEEE/CVF Conference on Computer Vision and Pattern
  Recognition (CVPR)}, pages 9300--9308, 2019.

\bibitem{deform-detr}
X. Zhu, Weijie Su, Lewei Lu, Bin Li, X. Wang, and Jifeng Dai.
\newblock Deformable detr: Deformable transformers for end-to-end object
  detection.
\newblock {\em ArXiv}, abs/2010.04159, 2020.

\bibitem{edgebox}
C.~Lawrence Zitnick and Piotr Doll\'ar.
\newblock Edge boxes: Locating object proposals from edges.
\newblock In {\em ECCV}, 2014.

\bibitem{self-training}
Barret Zoph, Golnaz Ghiasi, Tsung{-}Yi Lin, Yin Cui, Hanxiao Liu, Ekin~Dogus
  Cubuk, and Quoc Le.
\newblock Rethinking pre-training and self-training.
\newblock In Hugo Larochelle, Marc'Aurelio Ranzato, Raia Hadsell,
  Maria{-}Florina Balcan, and Hsuan{-}Tien Lin, editors, {\em Advances in
  Neural Information Processing Systems 33: Annual Conference on Neural
  Information Processing Systems 2020, NeurIPS 2020, December 6-12, 2020,
  virtual}, 2020.

\bibitem{od-20y}
Zhengxia Zou, Z. Shi, Yuhong Guo, and Jieping Ye.
\newblock Object detection in 20 years: A survey.
\newblock {\em ArXiv}, abs/1905.05055, 2019.

\end{thebibliography}
}

\end{document}